\documentclass[letterpaper, 10 pt, conference]{ieeeconf}  
\IEEEoverridecommandlockouts                            

\usepackage{amsmath,amssymb,amsfonts}
\usepackage{graphicx}
\usepackage{textcomp}
\usepackage{xcolor}
\usepackage{subcaption}
\usepackage{booktabs}
\usepackage{tabularx}
\usepackage{array}
\usepackage{xurl}
\usepackage{fancyhdr}
\usepackage{leftindex}

\fancypagestyle{firstpage}{%
  \lhead{Accepted to IEEE Robotics and Automation Letters (RA-L)}
  \rhead{}
}
\newtheorem{lemma}{\textbf{Lemma}}
\newtheorem{remark}{\textbf{Remark}}
\newtheorem{proposition}{\textbf{Proposition}}
\newtheorem{definition}{\textbf{Definition}}

\title{\LARGE \bf
Contact-rich $\SE$-Equivariant Robot Manipulation Task Learning via Geometric Impedance Control
}

\author{Joohwan Seo\textsuperscript{1}, Nikhil P. S. Prakash\textsuperscript{1}, Xiang Zhang\textsuperscript{1}, Changhao Wang\textsuperscript{1}, \\
Jongeun Choi\textsuperscript{1,2}, Masayoshi Tomizuka\textsuperscript{1}, and Roberto Horowitz\textsuperscript{1}
\thanks{This research is partially funded by (1) the Tsinghua-Berkeley Shenzhen Institute (TBSI) phase II and (2) the Hong Kong Center for Construction Robotics Limited (HKCRC). Jongeun Choi was supported by the National Research Foundation of Korea (NRF) grant funded by the Korea government (MSIT). (No.RS-2023-00221762 and No.2021R1A2B5B01002620)}
\thanks{\textsuperscript{1} UC Berkeley, Department of Mechanical Engineering. Emails: {\tt \small \{joohwan\_seo, nikhilps, xiang\_zhang\_98, changhaowang, tomizuka, horowitz\} @berkeley.edu}}
\thanks{\textsuperscript{2} Yonsei University, School of Mechanical Engineering. Email: {\tt \small jongeunchoi@yonsei.ac.kr}%
}}

\definecolor{color0}{rgb}{0.8235,0,0} 
\definecolor{color1}{rgb}{0.07843,0.549,0.07843} 
\definecolor{color2}{rgb}{0,0,1} 
\definecolor{color3}{rgb}{1,0.5137,0.4824} 
\definecolor{color4}{rgb}{0.5098,0.8588,0.1961} 
\definecolor{color5}{rgb}{0.4314,0.7451,0.9804} 
\definecolor{color6}{rgb}{0.3451,0.3451,0.3451} 
\definecolor{color7}{rgb}{0.6863,0.6863,0.6863} 
\definecolor{color8}{rgb}{0,0,0} 
\definecolor{color9}{rgb}{0.75,0.25,0} 
\definecolor{color10}{rgb}{0,0.8,0} 
\definecolor{color11}{rgb}{0.44, 0, 0.8 } 

\newcommand{\so}{so(3)}
\newcommand{\se}{se(3)}
\newcommand{\SO}{SO(3)}
\newcommand{\SE}{SE(3)}

\newcommand{\Ad}{\text{Ad}}

\newcommand{\eg}{e_{_G}}
\newcommand{\ec}{e_{_C}}
\newcommand{\fg}{f_{_G}}

\newcommand{\pvec}{p}
\newcommand{\pvecd}{p_d}

\newcommand{\ev}{e_V}

\newcommand{\g}{g}

\begin{document}

\maketitle

\begin{abstract}
This paper presents a differential geometric control approach that leverages $\SE$ group invariance and equivariance to increase transferability in learning robot manipulation tasks that involve interaction with the environment. Specifically, we employ a control law and a learning representation framework that remain invariant under arbitrary $\SE$ transformations of the manipulation task definition. Furthermore, the control law and learning representation framework are shown to be $\SE$ equivariant when represented relative to the spatial frame.
The proposed approach is based on utilizing a recently presented geometric impedance control (GIC) combined with a learning variable impedance control framework, where the gain scheduling policy is trained in a supervised learning fashion from expert demonstrations. A geometrically consistent error vector (GCEV) is fed to a neural network to achieve a gain scheduling policy that remains invariant to arbitrary translation and rotations. A comparison of our proposed control and learning framework with a well-known Cartesian space learning impedance control, equipped with a Cartesian error vector-based gain scheduling policy, confirms the significantly superior learning transferability of our proposed approach. 
A hardware implementation on a peg-in-hole task is conducted to validate the learning transferability and feasibility of the proposed approach. The simulation and hardware experiment video is posted: \url{https://sites.google.com/berkeley.edu/equivariant-task-learning/home}
\end{abstract}


\thispagestyle{firstpage}
\section{Introduction}
\vspace{-6pt}
Learning has become a prevalent method for robots to acquire skills in automated manipulation tasks \cite{ravichandar2020recent}.
Most learning-based approaches are formulated using a Cartesian frame to represent the end-effector workspace. However,  Cartesian-based learning formulations lack learning transferability in that trained policies cannot be directly transferred to arbitrarily translated/rotated task descriptions unless extensive additional training is conducted \cite{zhang2021learning, beltran2020variable}.

From the differential geometric perspective, the poor transferability of trained policies can be directly attributed to the absence of $\SE$ group equivariance in the Cartesian-based learning framework. For instance, a group transformation (e.g. translation/rotation) of the task definition does not result in a corresponding transformation of trained policy and task execution. As a consequence, group-transformed tasks have to be relearned, requiring a significant increase in training episodes and learning resources. 

Recently, in the field of geometric deep learning (GDL) in computer vision applications \cite{cohen2016group, bekkers2018roto}, the symmetry inherent to a group structure within a domain has been exploited and integrated into neural network models to enhance learning transferability and robustness to untrained data, thereby improving sample efficiency. 
The key properties exploited by GDL are group invariance and equivariance.
Invariance refers to a property of a map whose output remains unchanged when a group action transforms its input, while equivariance implies that a map's output is transformed by the representation of the same group action when its input is transformed by the group action.

Robotic manipulator workspaces have a group structure, which is frequently represented by the Special Euclidean group $\SE$ \cite{murray1994mathematical}. In our previous work \cite{seo2023geometric}, we introduced a geometric impedance control (GIC) for robot manipulators that fully incorporates the geometric structure of $\SE$. GIC leverages the left-invariance of the distance metric and potential function in $\SE$ to design a control law expressed in the body-frame coordinate system.

In this paper, we leverage $\SE$ group invariance and equivariance, extensively studied in GDL, to enhance learning transferability and sampling efficiency in contact-rich robotic manipulation tasks. A Peg-in-Hole (PiH) task is used as a testbed for evaluation since it is sensitive to $\SE$ transformations (i.e. translation/rotation) and involves contact-rich robot interaction with the environment.
A geometric learning variable impedance control is presented that utilizes the GIC in \cite{seo2023geometric} and incorporates a learning variable impedance control framework, where the gain scheduling policy is trained in a supervised learning fashion from expert demonstrations. A  geometrically consistent error vector (GCEV) is fed to a neural network to achieve a gain scheduling policy that remains invariant under arbitrary $\SE$ transformations.
The main contributions of our paper are:
\begin{enumerate}
    \item We propose key components for learning transferability in robotic manipulation tasks under dynamic feedback control law: left-invariance and a control law formulated in the body-frame coordinate system.
    \item We provide a theoretical justification for our proposed approach and demonstrate how learning transferability can be achieved through these components.
    \item We show that the proposed approach is equivariant when described in the spatial frame, as the trained policy and control law are invariant in the body frame. 
    \item We validate the feasibility of the proposed approach through a hardware experiment, including a workflow for data collection and policy training.
\end{enumerate}
We have also trained a policy via reinforcement learning and applied the proposed approach to surface wiping tasks. However, these results are not included due to the page length constraints and can be found in the project website. 

\vspace{-6pt}
\section{Related Works}
\vspace{-4pt}
\subsection{Geometric Deep Learning for Robotics Problem}
\vspace{-4pt}
Geometric Deep Learning (GDL) has found success in image-based data, particularly in medical fields \cite{bekkers2018roto}, and has also been applied in robotics \cite{zeng2021transporter,simeonov2022neural,ryu2022equivariant}. These geometric approaches leverage group structures' invariance and equivariance. For instance, SE(2) equivariance in \cite{zeng2021transporter} for image-based inputs, while SE(3) equivariance is explored in \cite{simeonov2022neural,ryu2022equivariant} for point cloud input data. \cite{simeonov2022neural} introduces a descriptor field with SE(3) equivariance, improving performance for out-of-distribution inputs, with further enhancement in \cite{ryu2022equivariant, ryu2023diffusion, kim2023robotic}.

Equivariant structures have been pursued in other robotics applications as well \cite{pan2023tax,ha2022flingbot,kim2023se}. In \cite{pan2023tax}, the translational equivariance is achieved via soft correspondence. \cite{ha2022flingbot} attains equivariance on the Euclidean group E(2) through data augmentation, which is data-inefficient. \cite{kim2023se} leverages SE(2) equivariance for learning tabletop object manipulation but only within the SE(2) space, without extending equivariance to control inputs.

Equivariant RL approaches have also been proposed in \cite{van2020mdp,wang2022equivariant,wang2022so2} to enhance RL's sample efficiency by leveraging symmetries in the Markov Decision Process (MDP). However, these equivariant RL approaches are generally applicable to simple scenarios without consideration of dynamics or force interaction \cite{van2020mdp} or are limited to SE(2) or SO(2) scenarios with actions restricted to displacements \cite{wang2022equivariant,wang2022so2}.

To date, the application of GDL in robotics has mainly focused on point cloud or image inputs and simpler task types without force interactions. Additionally, while control methods remain prevalent in robotics research, the equivariance properties of dynamic controls have yet to be fully explored, emphasizing the growing need to integrate GDL concepts with dynamic robot controllers.
\vspace{-6pt}
\subsection{Peg-in-Hole task and Learning Variable Impedance Control}
\vspace{-4pt}
The Peg-in-Hole (PiH) task is a benchmark problem for force-controlled robotic manipulation tasks \cite{inoue2017deep}. A widely utilized approach to solve a PiH task is the variable impedance control \cite{beltran2020variable, kozlovsky2022reinforcement, zhang2021learning}, where the controller's impedance gains change depending on the states. These approaches, however, do not take into account the geometric structures of the manipulator. As a result, it is reported in \cite{zhang2021learning} that the success rate of a trained policy dropped significantly under $\SE$ transformations, i.e., when the peg position is tilted relative to the orientation used for training. To deal with this issue, a typical approach is to adopt the domain randomization technique as in \cite{beltran2020variable}, where both the initial and goal poses of the end-effector are randomized during the training. 
In contrast, our proposed approach achieves robustness to out-of-distribution (OOD) goal poses and learning transferability without randomizing goal poses during the training stage.
We note that the proposed approach is not constrained to the PiH task, but it can be extended to other manipulation tasks with force interaction, such as surface wiping or pick-and-place. 
\vspace{-6pt}
\section{Preliminaries} \label{sec:3}
\vspace{-4pt}
\subsection{Lie Groups and Manipulator Dynamics}
\vspace{-4pt}
The configuration of the manipulator's end-effector can be defined by its position and orientation, and the configuration manifold lies in the Special Euclidean group $\SE$. We can represent the end-effector's configuration frame $\{e\}$ using the following homogeneous matrix $g_{se}$, to fixed a (inertial) spatial frame $\{s\}$,  as follows: 
\begin{equation}\label{eq:homo}
    g_{se} = \begin{bmatrix}
    R & \pvec \\ 0 & 1
    \end{bmatrix}\!\in\!\SE,
\end{equation}
where $R$ is a rotation matrix and $R \in \SO$, and $p \in \mathbb{R}^3$. We will drop the subscript $s$ since the spatial coordinate frame can be considered as an identity without loss of generality. In addition, we also drop the subscript $e$ for the current configuration of the end-effector for notational compactness unless specified, i.e., $g_{se} = g$. 
We also use $g = (R,p)$ for notational compactness. 

The Lie algebra of $\SE$, $\se$ can be represented by
\begin{align*}
    \hat{\xi} \!=\! \begin{bmatrix}
    \hat{\omega} & v \\ 0 & 0
    \end{bmatrix}\!\in\!\se, \; \forall \xi \!=\! \begin{bmatrix} v\\ \omega \end{bmatrix}\!\in\!\mathbb{R}^6, \; v,\omega\!\in\!\mathbb{R}^3, \;\hat{\omega} \!\in\! \so.
\end{align*}
For the details of the Lie group for robotic manipulators, we refer to \cite{murray1994mathematical, lynch2017modern}. Note also that we utilize the standard hat-map and vee-map notations as defined in \cite{seo2023geometric}.


The velocity of the end-effector relative to its body frame,
$\left.V^b \in \mathbb{R}^{6}\right.$, can be calculated by:
\begin{equation}\begin{aligned}
    V^b = \begin{bmatrix}
    v^b \\ \omega^b
    \end{bmatrix} = 
    (g^{-1} \dot{g})^\vee,
\end{aligned}\end{equation}
i.e., $\dot{g} = g \hat{V}^b$.
The velocity $V^b$ can also be computed using the body Jacobian matrix $J_b(q)$ via $V^b = J_b(q) \dot{q}$.
For the details about $J_b(q)$, we refer to Chap 5.1 of \cite{lynch2017modern}.

The dynamic equations of motion for rigid-link robotic manipulators are given by:
\begin{equation}
    \begin{split} \label{eq:robot_dynamics}
        M(q)\ddot{q} + C(q,\dot{q})\dot{q} + G(q) = T + T_e,
    \end{split}
\end{equation}
where $M(q)\!\in\!\mathbb{R}^{n\times n}$ is the symmetric positive definite inertia matrix, $C(q,\dot{q})\!\in\!\mathbb{R}^{n \times n}$ is a Coriolis matrix, $G(q)\!\in\!\mathbb{R}^n$ is a moment term due to gravity, $T\!\in\!\mathbb{R}^n$ is a control input, and $T_e\!\in\!\mathbb{R}^n$ is an external disturbance. As in \cite{seo2023geometric}, the manipulator dynamics in operational space formulation \cite{khatib1987unified}, using the body-frame velocity, is represented as follows:
\begin{align} \label{eq:robot_dynamics_eef}
    \tilde{M}(q)\dot{V}^b &+ \tilde{C}(q,\dot{q})V^b + \tilde{G}(q) = \tilde{T} + \tilde{T}_e, \text{ where}
\end{align}
{\small
\begin{align*}    
    \tilde{M}(q)&= J_b(q)^{-T} M(q) J_b(q)^{-1}, \nonumber\\
    \tilde{C}(q,\dot{q})&=J_b(q)^{-T}(C(q,\dot{q}) \!-\! M(q) J_b(q)^{-1}\dot{J})J_b(q)^{-1}, \nonumber\\
    \tilde{G}(q)&= J_b(q)^{-T} G(q),\; \tilde{T}= J_b(q)^{-T} T,\; \tilde{T}_e= J_b(q)^{-T} T_e \nonumber,
\end{align*}}
where $A^{-T} = {(A^{-1})}^T$. We will denote $\tilde{M}(q)$ as $\tilde{M}$, $\tilde{C}(q,\dot{q})$ as $\tilde{C}$ and $\tilde{G}(q)$ as $\tilde{G}$ for the rest of the paper.
\vspace{-6pt}
\subsection{Geometric Impedance Control Law} 
\vspace{-4pt}
We employ the geometric impedance control (GIC) law proposed in \cite{seo2023geometric}. We first note that $g$ denotes the current configuration matrix, with $p$ and $R$ in \eqref{eq:homo} representing the current position and rotation matrix, i.e., $g = (R,p)$. In a similar way, $g_d = (R_d,p_d)$ denotes the desired current configuration matrix.
In a nutshell, a GIC control law $\tilde{T}\in se^*(3)$ in the wrench is given by
\begin{equation} \label{eq:control_law}
    \tilde{T} = \tilde{M} \dot{V}_{d}^* + \tilde{C} V_d^* + \tilde{G} - \fg - K_d \ev.
\end{equation}
where $\tilde{C}$, and $\tilde{G}$ are matrices in the operational space formulation \eqref{eq:robot_dynamics_eef}, $K_d$ is symmetric positive definite damping matrix, $\fg$ is the elastic generalized force in $se^*(3)$ and $\ev$ is a velocity error vector, both of them described on the body-frame, which will be described subsequently. 
The elastic geometric wrench (force) $\fg(g,g_d)$ is given by
\begin{equation} \label{eq:geometric_elastic}
    \fg(g,g_d) = \begin{bmatrix}
    f_p \\ f_R
    \end{bmatrix} =\begin{bmatrix}
    R^T R_d K_p R_d^T (\pvec - \pvecd)\\
    (K_R R_d^T R - R^T R_d K_R)^\vee
    \end{bmatrix},
\end{equation}
where $K_p$ and $K_R$ denote symmetric positive definite stiffness matrices in translation and rotation, respectively. \\
%
The current and desired velocity vectors cannot be directly compared, as they lie on different tangent spaces. Therefore, we utilize a vector translation map (Adjoint map) to first translate the desired velocity to the tangent space of the current velocity \cite{bullo1999tracking}. 
The error vector $\ev$ is defined by
\begin{equation}
    \begin{aligned} \label{eq:velocity_error}
        \ev &= V^b - V_d^*  = [e_v^T, e_\omega^T]^T\\
        V_d^* &= \text{Ad}_{\g_{ed}}V_d^b, \text{ with } \text{Ad}_{\g_{bd}} = \begin{bmatrix}
           R_{ed} & \hat{\pvec}_{ed} R_{ed} \\ 0 & R_{ed}
        \end{bmatrix},
    \end{aligned}    
\end{equation}
where $V_d^b$ is a desired velocity in the desired body frame, $\text{Ad}_{\g_{ed}}: \mathbb{R}^6 \to \mathbb{R}^6$ is an Adjoint map, $R_{ed}\! =\! R^T R_d$, $p_{ed} \!=\!-R^T(p\! -\! p_d)$, and $V_{d}^*$ is a translated desired velocity on the configuration body-frame.

For the PiH task, we assumed that the final desired configuration $g_d$ is given and is time-invariant, i.e., $\dot{g}_d(t) = 0$. Thus, the controller law \eqref{eq:control_law} is modified as
\begin{equation}\label{eq:control_law_mod}
    \tilde{T} = - \fg - K_d V^b + \tilde{G},
\end{equation}
which can be interpreted as a  PD control together with gravity compensation. For more details on the GIC, such as derivation and stability properties, we refer to \cite{seo2023geometric} and its references \cite{lee2010geometric, bullo1999tracking}.
Here, we define a geometrically consistent error vector (GCEV) $\eg$, which will be utilized in the learning impedance gains later in this paper. The GCEV $\eg(g,g_d)$ is defined as follows:
\begin{equation} \label{eq:eg}
    e_{_G}(g,g_d) = \begin{bmatrix}
    e_p \\ e_R
    \end{bmatrix} = \begin{bmatrix}
    R^T(\pvec - \pvecd) \\
    (R_d^T R - R^T R_d)^\vee 
    \end{bmatrix} \in \mathbb{R}^6.
\end{equation}
Finally, we note that all the vectors/wrench, such as $\eg$, $\ev$, and $\fg$, are described in the body-frame coordinate unless otherwise specified.
\vspace{-6pt}
\subsection{Cartesian space Impedance Control}
\vspace{-4pt}
As a benchmark approach for the proposed GIC, we also briefly introduce a Cartesian space Impedance Controller (CIC), which is a currently standard method for impedance control. In the operational space formulation, correctly representing the rotational dynamics has always received significant interest \cite{caccavale1999six}. In particular, a positional Cartesian error vector $e_{_C}$ widely utilized in CIC can be defined in the following way \cite{robosuite2020, ochoa2021impedance, shaw2022rmps}.
\begin{align}
     e_{_C} &= [
     (e_{_{C,p}})^T, \; (e_{_{C,R}})^T
     ]^T, \quad \text{where}\\
    e_{_{C,p}} &= p \!-\! p_d, \quad e_{_{C,R}} = (r_{d_1}\! \times\! r_1 + r_{d_2}\! \times \! r_2 + r_{d_3}\! \times\! r_3), \nonumber
\end{align}
with $R = [r_1, r_2, r_3]$ and $R_d = [r_{d_1}, r_{d_2}, r_{d_3}]$. 
Utilizing the positional error vector in Cartesian space and considering a fair comparison with the GIC formulation, we will utilize the following CIC formulation for the PiH task. 
\begin{equation} \label{eq:conventional_impedance}
    \begin{split}
        \tilde{T}_{_C} =  - K_{_C} e_{_C} - K_{d_C} V^s + \tilde{G}_{_C} ,
    \end{split}
\end{equation}
swhere $\tilde{G}_{_C}$ can be obtained by replacing $J_b$ by $J_s$ in \eqref{eq:robot_dynamics_eef}, $K_{d_C}$ is a damping matrix for the CIC, and $K_{_C} = \text{blkdiag}(K_{_{C,p}}, K_{_{C,R}})$ with $K_{_{C,p}}$ and $K_{_{C,R}}$ are translational and rotational stiffness matrices, respectively. In addition, $J_s$ denotes a spatial frame Jacobian matrix and $V^s = J_s \dot{q}$. To implement control law \eqref{eq:control_law_mod} and \eqref{eq:conventional_impedance}, the wrenches should first be converted to joint torque $T$ in \eqref{eq:robot_dynamics} by multiplying corresponding Jacobian matrices, i.e., $T = J_b^T\tilde{T}$ and $T = J_s^T\tilde{T}_{_C}$, respectively.

We highlight the differences between the GIC and the CIC in the following remark:
\begin{remark} \textbf{Differences between GIC and CIC}\\
    The main differences between the GIC and the CIC are twofold: \\
    1. CIC deals with translational and rotational dynamics separately, but GIC deals with translational and rotational dynamics as a unified entity, using the geometric elastic wrench $\fg(g,g_d)$ in \eqref{eq:geometric_elastic}. \\
    2. GIC utilizes a body-frame Jacobian $J_b$ while CIC utilizes a spatial-frame Jacobian $J_s$. Therefore, we can interpret that GIC is formulated on the body frame coordinate attached to the end-effector, while CIC is formulated relative to the spatial frame.
\end{remark}
\vspace{-6pt}
\section{Problem Definition and Solution Approach}
\vspace{-4pt}
\subsection{Problem Setup}
\vspace{-4pt}
\subsubsection{Overview}
Our ultimate goal is to provide a learning-based solution to solve a peg-in-hole task, a classic representative of contact-rich force-based robotic manipulation tasks. 
We will address this problem in the framework of learning variable impedance control, where the gain scheduling policy of the impedance control laws is trained using learning algorithms. In particular, behavior cloning (BC) from expert demonstrations is utilized in a supervised learning fashion to obtain the gain scheduling policy. 

To achieve this, we introduce a gain scheduling policy parameterized by a simple neural network. This neural network takes as input the positional signals representing the current end-effector pose and the desired goal pose, such as $\eg$ and $\ec$. We consider the gain-scheduling neural network policy output as an action ($a_t$) and input positional signals as the states ($s_t$). Formally, 
\begin{equation}
    a_t = \mu_\theta(s_t), \quad (K_p, K_R)_t = h(a_t) = \pi_\theta(s_t),
\end{equation}
where $h$ denotes a mapping from the action signal to the gains, $K_p$ and $K_R$ are impedance gains defined in \eqref{eq:geometric_elastic} or \eqref{eq:conventional_impedance}, and $\theta$ denotes parameters of the neural network $\mu_\theta$. We employ a standard multi-layer perceptron (MLP) as a neural network. For the rest of the paper, we will call $\pi_\theta$ as a gain-scheduling policy and drop the subscript $t$ for the compactness of notation. 

To show the effectiveness of the geometric formulation, we will propose and compare two different approaches to learning variable impedance control: 
1. Selection of the control rule (GIC vs CIC), 2. Selection of the states $s_t$ (Geometric error vs Cartesian error).

\begin{figure*}[ht]
    \centering
    \begin{subfigure}{0.19\linewidth}
        \centering
        \includegraphics[width=0.99\linewidth]{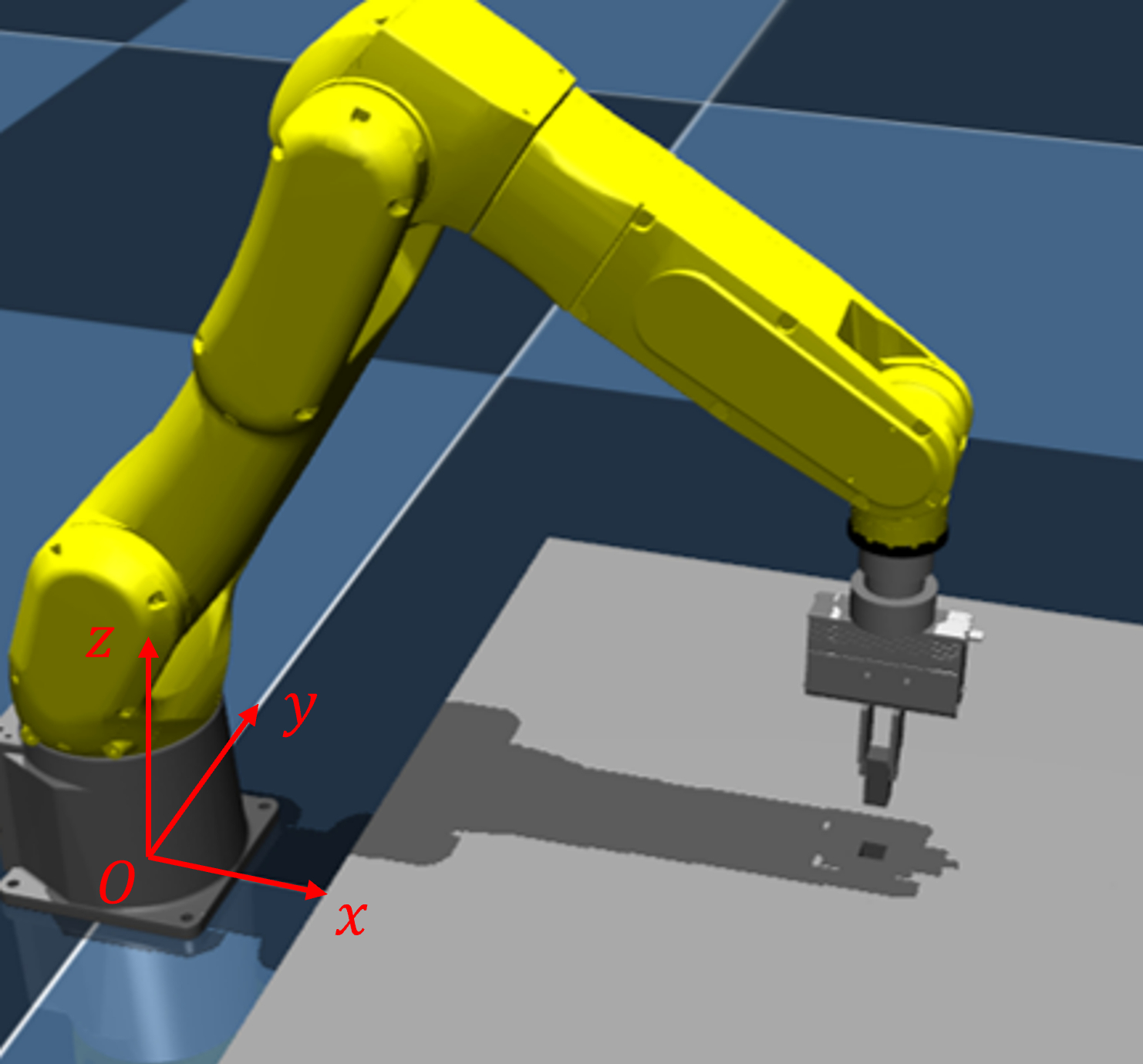}
        \caption{Default Case}
    \end{subfigure}
    \begin{subfigure}{0.19\linewidth}
        \centering
        \includegraphics[width=0.99\linewidth]{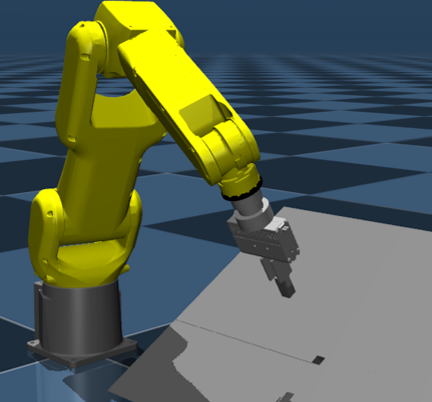}
        \caption{Case 1}
    \end{subfigure}
    \begin{subfigure}{0.19\linewidth}
        \centering
        \includegraphics[width=0.99\linewidth]{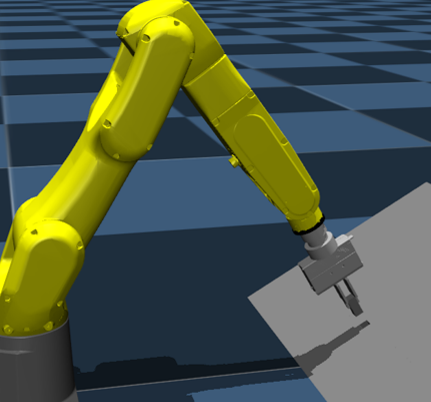}
        \caption{Case 2}
    \end{subfigure}
    \begin{subfigure}{0.19\linewidth}
        \centering
        \includegraphics[width=0.99\linewidth]{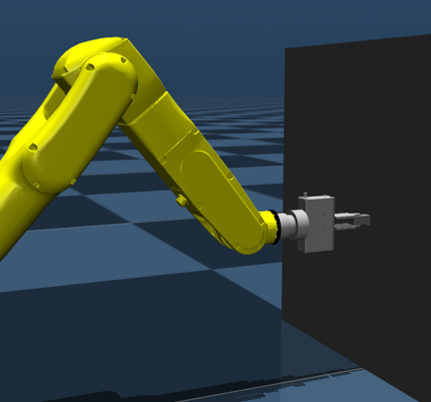}
        \caption{Case 3}
    \end{subfigure}
    \begin{subfigure}{0.19\linewidth}
        \centering
        \includegraphics[width=0.99\linewidth]{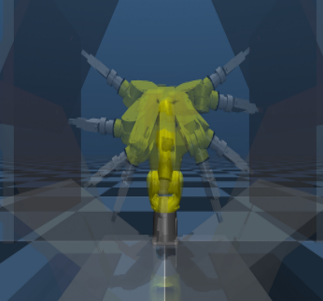}
        \caption{Case 4}
    \end{subfigure}
    \vspace{-5pt}
    \caption{Robot performing a peg into a hole insertion task in different scenarios for testing learning transferability. The data is collected, and the policy is trained only in performing the task shown in (a). The trained policy is then tested on the tasks shown in (b)-(e), where the insertion hole is translated in different orientations. (b): tilted in $+x$ direction for $30^\circ$ (c): tilted in $-y$ direction for $30^\circ$ (d): tilted in $-y$ direction for $90^\circ$ (e): tilted in $+x$ direction in arbitrary angle $\phi \in [-135^\circ, 135^\circ]$. The coordinate frame is attached to the first figure. \vspace{-15pt}}
    \label{fig:pih_scenarios}
\end{figure*}

The performances are evaluated on four main scenarios in Fig.~\ref{fig:pih_scenarios}. The gain scheduling policy $\mu_\theta$ is trained only in the default scenario (Fig.~\ref{fig:pih_scenarios}(a)). The trained policy is then tested in the other scenarios (Fig.~\ref{fig:pih_scenarios}s(b)-(d)) to evaluate its zero-shot transferability and robustness to OOD data. All simulations are conducted in the Mujoco simulation environment \cite{todorov2012mujoco} for environment setup and the Berkeley RL kit \cite{BerkeleyRLkit} for the RL training. The GitHub repository of this project is published in \url{https://github.com/Joohwan-Seo/GIC_Learning_public}.
\subsubsection{Action Mapping to Gains}
The mapping from the actions to the impedance gains $h(a)$ is defined here.
To map the actions from policy to impedance gains, we first consider diagonal components of the matrix gains $K_p$ and $K_R$ as follows:
\begin{equation*}
    K_p = \text{diag}([k_{p_1}, k_{p_2}, k_{p_3}]), \; K_R = \text{diag}([k_{r_1}, k_{r_2}, k_{r_3}]).
\end{equation*}
The damping matrix $K_d$ is fixed as 
\begin{equation}
    K_d = 8 \cdot \text{diag}([k_{p_1}, k_{p_2}, k_{p_3}, k_{r_1}, k_{r_2}, k_{r_3}])^{0.5}
\end{equation}

For the task shown in Fig.~\ref{fig:pih_scenarios}(a), a symmetric structure is considered in the action mapping. Specifically, we have the same action mapping in the $x$ and $y$ directions ($k_{p_1}$ and $k_{p_2}$) and a different mapping in the $z$ direction. In the rotational part, the action mapping for $k_{r_1}$, $k_{r_2}$, and $k_{r_3}$ is the same. The selected action mapping is as follows:
\begin{equation*}
    \begin{split}
        k_{p_i} &= 10^{a_i + 2.5}, \quad \text{for} \; i = 1,2, \quad k_{p_3} = 10^{1.5 \cdot a_3 +2.0},\\
        k_{r_j} &= 10^{0.6 \cdot a_j + 2.0}, \quad \text{for} \; j = 1,2,3,
    \end{split}
\end{equation*}
where $a$ denotes an action, $a  = [a_1, a_2, \cdots, a_6]^T \in [-1,1]\times \cdots \times[-1,1] \subset \mathbb{R}^6$. 
\vspace{-6pt}
\subsection{Solution Approach}
\vspace{-4pt}
In this subsection, we will first introduce our proposed approach. In what follows, the behavior cloning (BC) to obtain the gain-scheduling policy is introduced.
\subsubsection{Proposed Approach}
Our proposed approach utilizes GIC with a learning impedance gain scheduling policy, where the input to the neural network $s$ is the GCEV $e_g(g,g_d)$ defined in \eqref{eq:eg}.
The action from the gain scheduling policy then becomes $(K_p, K_R) = \pi_\theta(\eg)$. The GIC control law \eqref{eq:geometric_elastic} equipped with gain scheduling policy $\pi_\theta(\eg)$ has a crucial property for learning transferability as shown in the following lemma.
\begin{lemma}\label{lem:left_invariance}
    \textbf{Left-invariance of the GCEV \eqref{eq:eg} and the elastic wrench \eqref{eq:geometric_elastic}}\\
    $\eg(g,g_d)$ and $\fg(g,g_d)$ are left-invariant to the arbitrary left-transformation $g_l$ in $\SE$.
\end{lemma}
\begin{proof}
    Let $g_l = (R_l, p_l)$. Then, the left-transformed homogeneous matrix $g_l g$ is calculated in the following way:
    \begin{equation*}
        g_l g = \begin{bmatrix}
            R_l & p_l \\ 0 & 1
        \end{bmatrix} \begin{bmatrix}
            R & p \\ 0 & 1
        \end{bmatrix} = \begin{bmatrix}
            R_l R & R_l p + p_l \\ 0 & 1
        \end{bmatrix}
    \end{equation*}
    Similarly, $g_l g_d = (R_lR_d, R_l p_d + p_l)$. The left-transformed GCEV is then
    \begin{equation}
        \begin{split}
            \eg(g_l g, g_l g_d) &= \begin{bmatrix}
            R^T R_l^T \left(R_l p + p_l - R_lp_d - p_l \right) \\
            \left((R_l R_d)^T R_l R - (R_l R)^T R_l R_d\right)^\vee
            \end{bmatrix} \\
            &= \begin{bmatrix}
                R^T (p - p_d) \\
                (R_d^T R - R^T R_d)^\vee 
            \end{bmatrix} = \eg(g, g_d)
        \end{split}        
    \end{equation}
    As a result, the gain scheduling policy $\pi_\theta(\eg)$ is also left-invariant, i.e., let $(K_p, K_R) (g,g_d) = (\pi_\theta\circ \eg)(g,g_d)$, then $(K_p, K_R) (g_lg, g_l g_d) = (K_p, K_R) (g,g_d)$.
    Similarly, the left-transformed elastic wrench is
    \begin{align}
        &\fg(g_lg,g_lg_d) \nonumber\\
        &= \begin{bmatrix}
            (R_l R)^T R_l R_d K_p^l (R_l R_d)^T (R_l p + p_l - R_l p_d - p_l)\\
            (K_R^l(R_l R_d)^TR_lR - (R_l R)^T R_l R_d K_R^l)^\vee
        \end{bmatrix} \nonumber \\
        &= \begin{bmatrix}
            R^T R_d K_p R_d(p - p_d)\\
            (K_R R_d R - R^T R_d K_R)^\vee
        \end{bmatrix} = \fg(g,g_d),
    \end{align}
    where we used notations $(K_p^l, K_R^l) = (K_p, K_R)(g_l g, g_l g_d)$ and $(K_p, K_R) = (K_p, K_R)(g,g_d)$ to avoid clutter.
This shows the left-invariance of $\eg(g,g_d)$ and $\fg(g,g_d)$.
\end{proof}
\textbf{Domain Randomization:} Domain randomization is a crucial technique in both BC and RL to enhance robustness by allowing the neural network to explore a broader range of state space \cite{tobin2017domain}. In conventional impedance gains learning problem \cite{beltran2020variable}, domain randomization is typically applied to both the initial and goal end-effector poses. However, under the proposed GIC framework, domain randomization is only necessary for the initial pose of the end-effector, as demonstrated in the following proposition.
\begin{proposition} \label{pro:randomization}
    For the learning variable impedance control problem based on GIC law \eqref{eq:control_law}, the following equation holds true:
    \begin{equation} \label{eq:domain_randomization}
        \begin{split}
            \fg(g, g_l g_d) = \fg(g_l^{-1}g, g_d)
        \end{split}
    \end{equation}
\end{proposition}
\begin{proof}
    By lemma~\ref{lem:left_invariance}, $\fg(g,g_d)$ is left-invariant to arbitrary $\SE$ transformation $g_l$ acting on the left, i.e., $\fg(g_l g, g_l g_d) = \fg (g,g_d)$. Then, \eqref{eq:domain_randomization} becomes
    \begin{equation}
        \fg(g,g_l g_d) = \fg(g_l g_l^{-1} g, g_l g_d) = \fg(g_l^{-1}g,g_d).
    \end{equation}
\end{proof}
The effects of Proposition~\ref{pro:randomization} on learning strategies are as follows. First, since the main driving force of \eqref{eq:control_law_mod} with gain scheduling policy is $\fg(g,g_d)$ in \eqref{eq:geometric_elastic}, we focus on the properties of $\fg(g,g_d)$.
The domain randomization on the target pose can be represented by 
$\fg(g, g_l g_d)$ where $g_l \in \SE$ is arbitrary. 
Then, the result of Proposition~\ref{pro:randomization} reads that
\begin{equation*}
    \fg(g, g_{l} g_d) = \fg(g_l^{-1} g , g_d)
\end{equation*}
Note that following the axioms of groups (Chap 2.1, \cite{murray1994mathematical}), $g_l^{-1}$ can be denoted by another group element $g_l'$ since $g_{l}$ is arbitrary. 
Finally, $\fg(g_l'g, g_d)$ means that the domain randomization on the target pose of the end-effector is identical to randomization on the initial pose.
Therefore, independent domain randomizations on both the target pose and the initial pose are not necessary during the training process. Only the initial pose relative to the target pose needs to be randomized, or vice versa, but not both.
This result is crucial in robotics applications where the cost of collecting data in the real world is substantial.
\subsubsection{Behavior Cloning (BC)}
We employed naive behavior cloning to directly estimate the impedance gain for the given input states in a supervised learning fashion from the expert's demonstration. The learning problem for behavior cloning can be defined as $
 \theta^* = \arg \min_\theta \tfrac{1}{N}\sum_{i=1}^N\|a_i - \mu_\theta(s_i)\|_2^2,$
where $N$ is the length of the dataset. We collect $300$ expert demonstration trajectories with $N\sim 450k$.
The BC policy is trained following the standard deep learning fashion - stochastic gradient descent on a sampled batch dataset with an Adams optimizer, learning rate schedule, and early stopping.

To collect the required dataset in the simulation environment, we develop a heuristic rule-based scripted expert policy. 
The main intuition is to use small $z$ gains and high gains in the $x$ and $y$ directions when the robot approaches the hole while using high $z$ gains during the insertion to overcome the friction between the peg and the hole. As a result, the robot first aligns the peg with the hole's axis and then gradually pushes the peg into the hole.  The expert's gains are selected by trial and error, depending on the error signal $\eg$.
During data collection for BC, small amounts of noise are added to the scripted expert policy to enhance robustness. 
\vspace{-6pt}
\section{Experiments and Discussions}
\vspace{-4pt}
\subsection{Behavior Cloning (BC)}
The results of the BC experiments are presented in Table.~\ref{table:policy_result}. Each task was tested 100 times, and the success cases were counted.
The BC policy trained with the GCEV and executed with the GIC (GIC+GCEV) successfully transferred the trained policy to the other tasks, without significant drop in the success rate. However, the BC policy trained with a CEV and executed with CIC (CIC+CEV) failed to transfer the trained policy, resulting in a dramatic decrease in the success rate. The reason for this difference in transferability can be attributed to the error vector representation. The relationship between left invariance and transferability is further explained in Remark~\ref{rem:1}.

%

\begin{figure}[t]
    \centering
    \begin{subfigure}{0.45\columnwidth}
        \centering
        \includegraphics[width=0.7\linewidth]{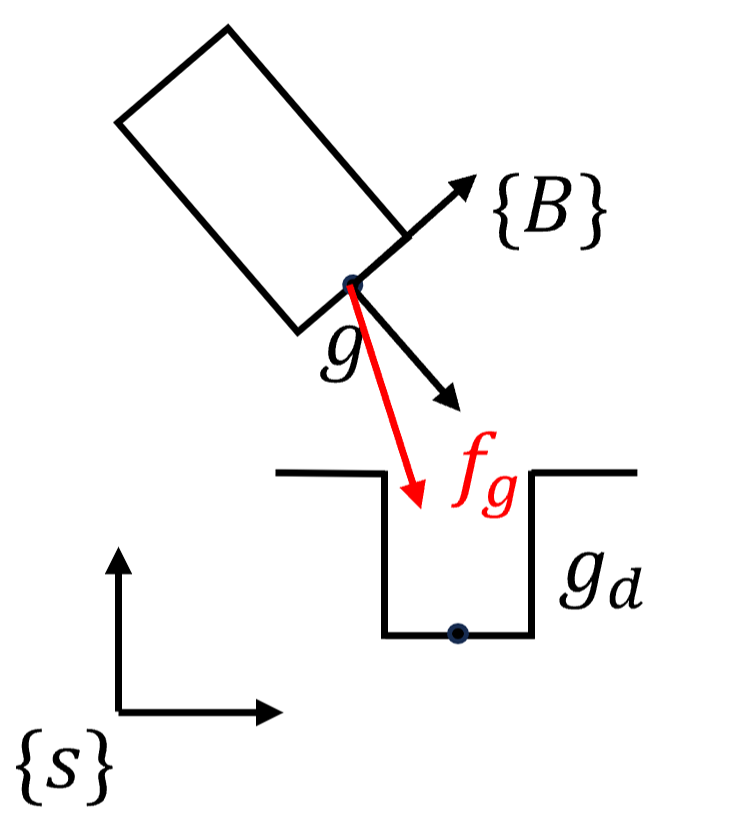}
    \end{subfigure}
    \begin{subfigure}{0.45\columnwidth}
        \centering
        \includegraphics[width=0.7\linewidth]{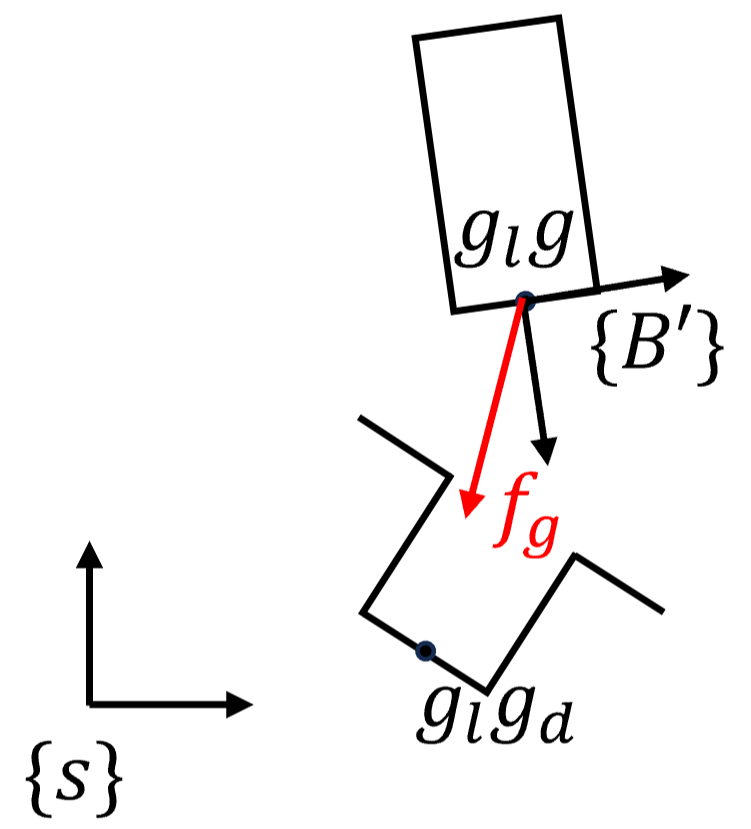}
    \end{subfigure}
    \vspace{-3pt}
    \caption{(Left) Peg and hole configurations are represented by $g$ and $g_d$, respectively. (Right) Peg and hole undergo left transformation via an action $g_l \in \SE$. The GCEV $\eg$ \eqref{eq:eg} and elastic force $\fg$ \eqref{eq:geometric_elastic} are invariant to the left transformation of $\SE$. $\{s\}$ represents spatial coordinate frame, while $\{B\}$ and $\{B'\}$ represent body coordinate frames.}
    \label{fig:pih_transformed}
\end{figure}

\begin{table}
    \setlength\doublerulesep{0.5pt}
    \renewcommand\tabularxcolumn[1]{m{#1}}
    \centering
    \caption{
    Success rates of the BC policies for the proposed and the benchmark approaches 
    (Tested $100$ times each, Values in Percentage \%)}
    \label{table:policy_result}
    \begin{tabularx}{\linewidth}{
     >{\raggedright\arraybackslash\hsize=2.6\hsize}X >{\centering\arraybackslash \hsize=0.6\hsize}X >{\centering\arraybackslash \hsize=0.6\hsize}X >{\centering\arraybackslash \hsize=0.6\hsize}X >{\centering\arraybackslash \hsize=0.6\hsize}X
    }
    \toprule[1pt]\midrule[0.3pt]
    Method & Default & Case 1 & Case 2 & Case 3\\
    \midrule
    BC Proposed (GIC+GCEV) & 100 & 99 & 95 & 100\\
    BC Benchmark (CIC+CEV) & 100 & 0 & 0 & 1\\
    \midrule
    BC Mixed 1 (GIC+CEV) & 99 & 54 & 49 & 27\\
    BC Mixed 2 (CIC+GCEV) & 100 & 0 & 0 & 0\\
    \midrule[0.3pt]\bottomrule[1pt]
    \end{tabularx}
    \vspace{-5pt}
\end{table}
\begin{remark} \textbf{Why does left-invariance matter?}\\
    \label{rem:1}
    %
    The left-invariance of the error vector implies that the chosen error vector is invariant to the selection of the coordinate system. In this paper, we interpret left-invariance in a slightly different manner. Consider the situation where the desired and current configurations are transformed through a left action of the SE(3) group, which corresponds to a change in the spatial coordinate frame -- See Fig.~\ref{fig:pih_transformed}. Due to the left-invariance of the GCEV \eqref{eq:eg}, the error vector remains unchanged in cases (a) and (b) in Fig.~\ref{fig:pih_transformed}. Therefore, from the perspective of the proposed approach, the task remains invariant to translational/rotational perturbations. In this perspective, the use of the left-invariant error vector $\eg$ can help address distributional shift or out-of-distribution issues, as the trained policy will consistently encounter the same input $\eg$.
\end{remark}
%
%
%
\vspace{-6pt}
\subsection{Left-Invariance is not enough}
The question arises whether training a policy based on GCEV or equivalent left-invariant features enables transferability to translational/rotational perturbations. To answer this question, we test different combinations of trained gain scheduling policies and control methods: BC policy with Cartesian error vector executed with GIC (GIC+CEV) and BC policy with a GCEV executed with CIC (CIC+GCEV). If left-invariance of the feature were the only factor for transferability, we would expect the BC policy with the Cartesian error vector executed with GIC (GIC+CEV) to not be transferable due to the lack of left-invariance in the Cartesian error vector. Conversely, the BC policy using a GCEV executed with CIC would be transferable.

However, the experimental results presented in $3\textsuperscript{rd}$ and $4\textsuperscript{th}$ rows of Table~\ref{table:policy_result} contradict the hypothesis. The GIC with a gain scheduling policy trained with the Cartesian error vector (GIC + CEV) showed some transferability in Case 1 and Case 2, with success rates near 50\%. However, in Case 3, with a tilt of 90 degrees, the success rate drops to 27\%
due to encountering an unexperienced state distribution. This suggests that a left-invariant gain scheduling policy alone is not enough for transferability.

\begin{figure}
    \centering
    \begin{subfigure}{0.49\columnwidth}
        \centering
        \includegraphics[width=0.90\linewidth]{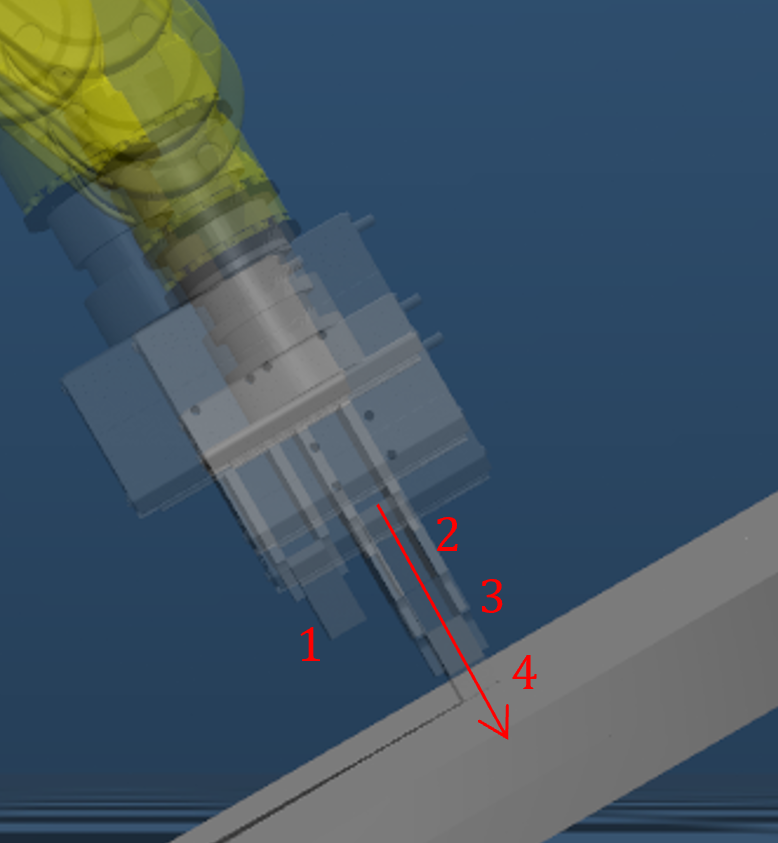}
    \end{subfigure}
    \begin{subfigure}{0.49\columnwidth}
        \centering
        \includegraphics[width=0.90\linewidth]{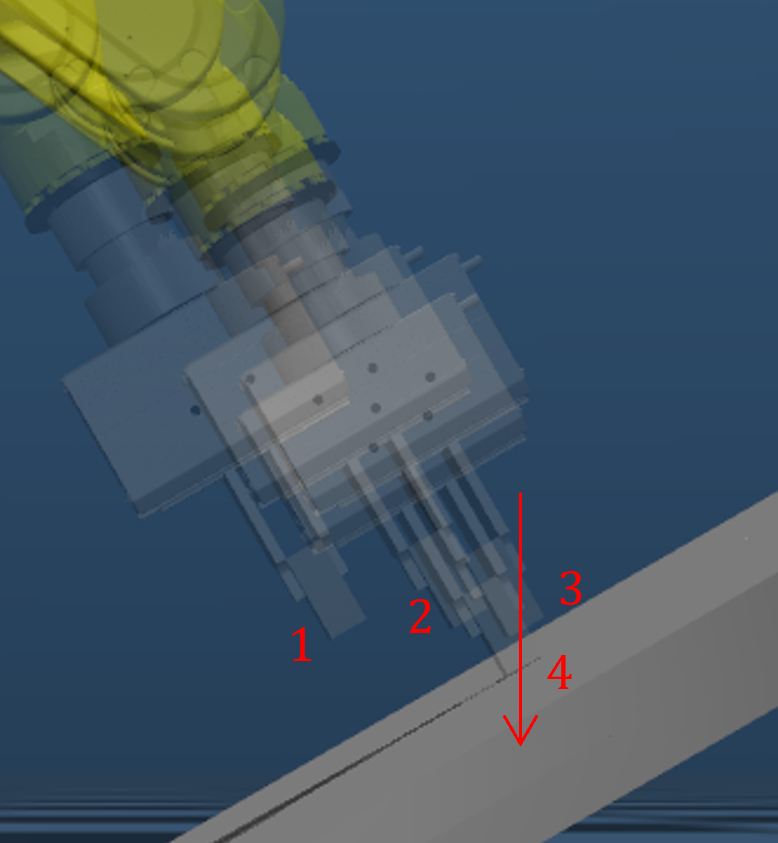}
    \end{subfigure}
    \vspace{-13pt}
    \caption{Time-snap plots of for PiH task executed by (Left) GIC+GCEV, and (Right) CIC+GCEV. The numbers in the figure denote the time snap index, and the red arrow denotes the insertion direction. The desired point is located at the tip of the arrow. While GIC successfully first aligns with the axis of the hole as expert policy, CIC first aligns with the Cartesian $z$ axis as it is trained in the default case (Fig.~\ref{fig:pih_scenarios}(a)).}
    \label{fig:comparison_GIC_CIC}
    \vspace{-5pt}
\end{figure}

On the other hand, the CIC using a GCEV produced results consistent with our hypothesis. Even though the gain scheduling policy for CIC was trained with a GCEV, the direction of the gains and resulting force direction are still represented in the Cartesian frame. Thus, the policy outputs the same gains it was trained on, but the resulting force direction is not suitable for tilted cases -- see in Fig.~\ref{fig:comparison_GIC_CIC}.



Consequently, it is concluded that the left-invariant gain scheduling policy alone is not enough, and a more fundamental factor is needed to address transferability - the direction of forces. Unlike CIC, the forces in GIC are defined in the body frame, resulting in the automatic change in the direction of forces (See Fig.~\ref{fig:pih_transformed}, $\fg$ on each case) -- which implies an \textbf{\emph{equivariance}} property. We state that the key to transferability lies in a control law represented in the body-frame coordinate and the left-invariant gain scheduling policy. The invariant gain scheduling policy is obtained from the neural network with a GCEV as input, while the left-invariant feedback control law is inherited from the structure of GIC. To establish a connection between our statement and the equivariance property, we first present the definition of equivariance.
\begin{definition} \label{def:equivariance}
    Consider a function $f: \mathcal{X} \to \mathcal{Y}$, i.e., $y = f(x)$ with $x \in \mathcal{X}$ and $y \in \mathcal{Y}$. The function $f$ is equivariant to the group $g$ if the following condition is satisfied \cite{cohen2016group}:
    \begin{equation}
        f(\rho^\mathcal{X} x) = \rho^\mathcal{Y} f(x)
    \end{equation}
    where $\rho^\mathcal{X}$ represents the action of group $g$ in the domain $\mathcal{X}$ and $\rho^\mathcal{Y}$ represents the action of group $g$ in the codomain $\mathcal{Y}$.
\end{definition}

Based on Definition~\ref{def:equivariance}, we propose the following proposition.
\begin{proposition} \label{pro:equivariance}
    The feedback terms in GIC law \eqref{eq:control_law} described in the body frame are equivariant if it is described in the spatial frame.
\end{proposition}
\begin{proof}
    Consider the feedback terms $\fg(g,g_d)$ \eqref{eq:geometric_elastic} and $e_V(g,g_d)$ \eqref{eq:velocity_error}, which are denoted on the body frame.
    See Fig.~\ref{fig:pih_transformed} for the coordinate systems. To show the equivariance property, we first show that the invariance on the body frame implies equivariance on the spatial frame. \\
    \indent Let $\fg^{s}(g,g_d)$ be $\fg(g,g_d)$ denoted in the spatial frame $\{s\}$. We note that $g$ and $g_d$ are described on the spatial frame $\{s\}$. Then, the left-transformed elastic force can be denoted by $\fg(g_lg,g_lg_d)$ on the transformed body frame $g_lg$ and is left-invariant by lemma~\ref{lem:left_invariance}, i.e., $\fg(g_lg,g_lg_d) = \fg(g,g_d)$. We now consider the left-transformed elastic force $\fg^{s}(g_lg,g_lg_d)$ with respect to the spatial frame $\{s\}$. The coordinates of the wrenches between the body and the spatial frame can be transformed by the following equations (see Ch. 2.5 of \cite{murray1994mathematical}):
    \begin{align}
        \fg^{s}(g,g_d) &= \Ad^T_{g^{\text{-}1}}\fg(g,g_d)\\
        \fg^{s}(g_lg,g_lg_d) &= \Ad^T_{(g_lg)^{\text{-}1}}\fg(g_lg,g_lg_d) = \Ad^T_{g^{\text{-}1}g_l^{\text{-}1}}\fg(g,g_d), \nonumber
    \end{align}
    where the Adjoint map is defined in \eqref{eq:velocity_error}.
    Therefore, the following equations hold:
    \begin{align}
        &\fg^{s}(g_lg,g_lg_d) = \Ad^T_{g^{\text{-}1}g_l^{\text{-1}}}(\Ad^T_{g^{\text{-}1}})^{\text{-}1}\fg^{s}(g,g_d) \\
        &\;\;= (\Ad_g \Ad_{g^{\text{-}1}g_l^{\text{-1}}})^T \fg^{s}(g,g_d)
        = \Ad^T_{g_l^{\text{-}1}}\fg^{s}(g,g_d), \nonumber
    \end{align}
    where we use a composition rule for the Adjoint map, $\Ad_{g_1}\Ad_{g_2} = \Ad_{g_1g_2}$, and an inverse property of the Adjoint map, $(\Ad^T_{g})^{-1} = \Ad^T_{g^{\text{-}1}}$. Here, the domain $\mathcal{X}$ in Def. \ref{def:equivariance} is $\SE$ ($g$ or $g_d$), and the representation $\rho^{\mathcal{X}}$ is $g_l$, while the codomain $\mathcal{Y}$ is $se^*(3)$ ($\fg^{s}(g,g_d)$ or $\fg^{s}(g_lg,g_lg_d)$) and the representation $\rho^\mathcal{Y}$ is $\Ad_{g_{l}^{\text{-}1}}^T$. Therefore, $\fg^s(g,g_d)$ is equivariant in $\SE$.\\
    \indent Similarly, to show the equivariance of $e_V(g,g_d)$ \eqref{eq:velocity_error} on the spatial frame, the invariance on the body frame is only needed. The body frame velocities $V^b$ and $V_d^b$ are invariant to the left transformation since it is defined on the body frame and the Adjoint map $\Ad_{g_{ed}}$ is invariant as the relative transformation matrix $g_{ed} = g^{-1}g_d$ is invariant to the left transformation. As a result, the feedback terms in GIC law \eqref{eq:control_law} are equivariant in $\SE$.
\end{proof}
Note that we only consider feedback terms in \eqref{eq:control_law} since the feedforward terms are just employed to cancel the manipulator dynamics, not affecting the closed-loop dynamics. 
A remark on the Proposition~\ref{pro:equivariance} is provided. 
\begin{remark} \textbf{Extensions to general force-based policy}\\
It is worth mentioning that our concept proposed in Proposition~\ref{pro:equivariance} can also be extended to general force-based policies. If a force-based policy is left-invariant in the body-frame, e.g., implemented using a neural network with left-invariant features as input, and described in the end-effector body frame, it will be guaranteed to be left-equivariant in the spatial frame. The GCEV \eqref{eq:eg} introduced in this paper is an example of such a left-invariant feature. 
\end{remark}


\vspace{-6pt}
\subsection{Demonstration of Sample Efficiency}
\vspace{-4pt}
In addition to its enhanced learning transferability, equivariant learning approaches offer a significant advantage in terms of increased sample efficiency. To evaluate whether our proposed method demonstrates improved sample efficiency, we trained a gain scheduling policy with CEV, making the GIC+CEV non-equivariant, while employing data augmentation, as illustrated in Case 4 of Figure~\ref{fig:pih_scenarios}(d).
To simplify the testing, we assumed that the goal pose is only rotated in the $+x$ axis. The dataset was collected across tilting angles ranging from $-90^\circ$ to $90^\circ$ in the $+x$ axis. As the state space region that the neural network needs to memorize expands, it becomes necessary to increase its size. 

The comparison results are summarized in Table~\ref{table:data_augmentation_test}. When tested within the training region, both GIC+GCEV and GIC+CEV with data augmentation showed near-perfect performance, with a slight advantage for GIC+GCEV. However, in OOD scenarios, the success rate of GIC+CEV with data augmentation dropped as it encountered OOD data inputs. Even with the goal pose tilting in only one direction, it required 8.6 times more data points and neural networks that were 2.3 times larger. It is important to note that if tilting occurs in different dimensions, the required samples and neural network size will increase in power scale. These results demonstrate the superior sample efficiency of the equivariant policy compared to the non-equivariant one.
%
{\color{blue}
\begin{table}
    \setlength\doublerulesep{0.5pt}
    \renewcommand\tabularxcolumn[1]{m{#1}}
    \centering
    \caption{Comparison between GIC+GCEV and GIC+CEV with data augmentation. Tested $100$ times for uniform randomly sampled tilting angles within range.}
    \label{table:data_augmentation_test}
    \begin{tabularx}{\linewidth}{
     >{\centering\arraybackslash\hsize=1.2\hsize}X >{\centering\arraybackslash \hsize=0.9\hsize}X >{\centering\arraybackslash \hsize=0.9\hsize}X
    }
    \toprule[1pt]\midrule[0.3pt]
     Default & GIC+GCEV & GIC+CEV w/ aug. \\
    \midrule
    Success Rate* & 99 \% & 93 \% \\
    Success Rate to OOD** & 96 \% & 43 \% \\
    Size of Dataset & 488{\small,}854 & 4{\small,}194{\small,}514\\
    Size of NN & $[128]\times3$ & $[128]\times7$ \\
    \midrule[0.3pt]\bottomrule[1pt]
    \end{tabularx}
{\raggedright *$\phi \in [-90^\circ,90^\circ]$, \\
**OOD: Out-of-distribution data, $\phi \in [-135^\circ,-90^\circ) \cup (90^\circ,135^\circ]$\par}
\end{table}}
\vspace{-6pt}
\section{Hardware Experiment}
\begin{figure}
    \centering
    \includegraphics[width=0.99\linewidth]{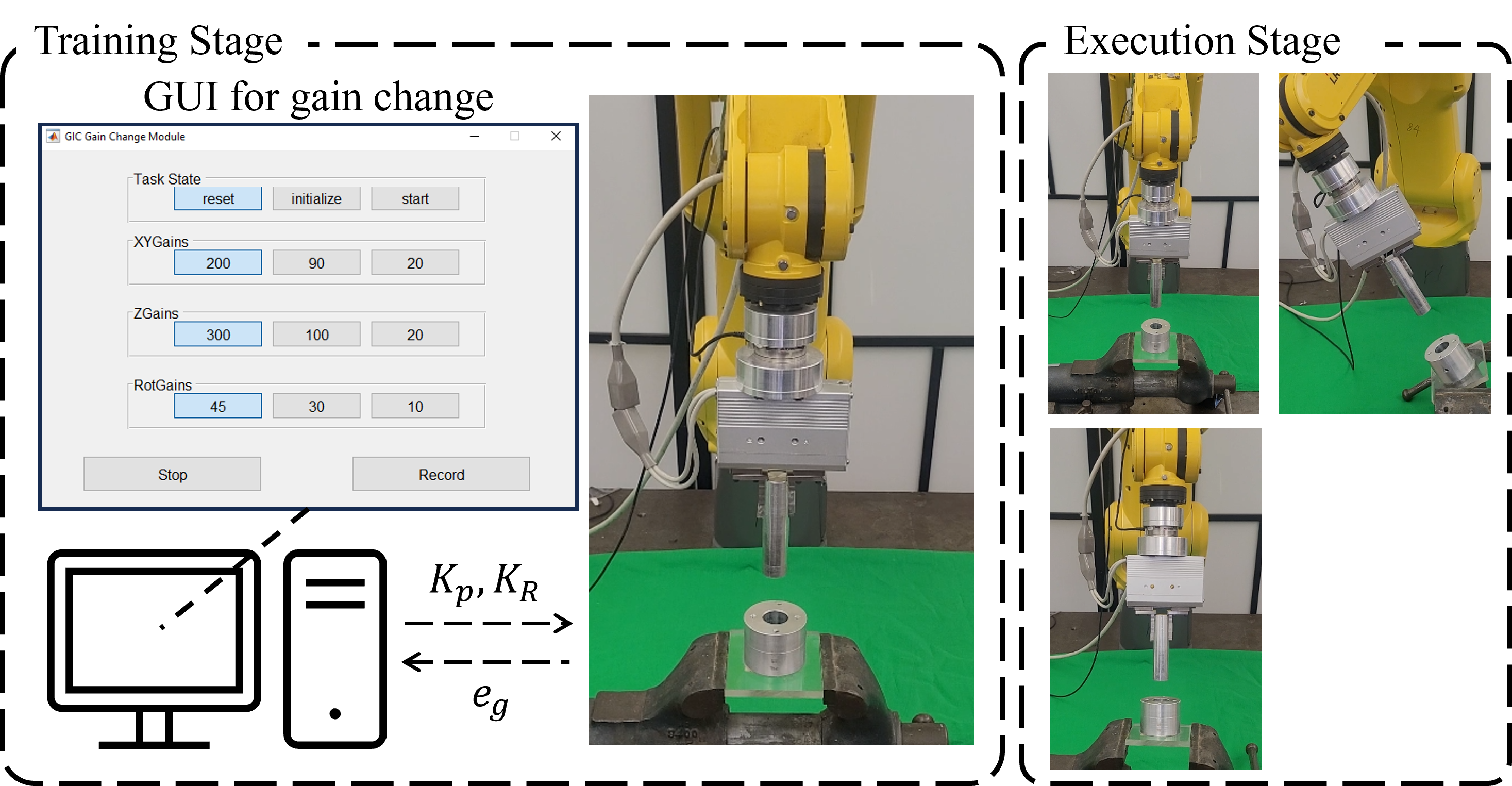}
    \vspace{-15pt}
    \caption{Real robot experiment description. \textbf{Training Stage}: Schematic for the data collection method is shown. The data is collected on the default case. \textbf{Execution Stage}: 
    The trained policy is then directly tested on the holes with different positions/orientations. Upper-left: Default Case. Upper-right: the hole is tilted $30^\circ$ in $+x$ axis (Case 1), Lower-left: $-22.5^\circ$ in $+y$ axis. (Case 2)}
    \label{fig:experiment_description}
\end{figure}

\vspace{-4pt}
To validate the proposed concept, we implemented our methods on the hardware robot; Fanuc 200iD/7L. The proposed approach is tested in a tight PiH task, where the clearance between the peg and hole is $0.05\mathrm{mm}.$ For ease of hardware implementation, we implement the GIC (geometric impedance control) as a Geometric Admittance Control (GAC) law. To define the admittance control version of the GIC (GAC), the desired closed-loop control system $M \dot{e}_V + K_d e_V + \fg(g,g_d) = \tilde{T}_e$ is first considered,
where $M$ is a fixed desired inertia matrix, $K_d$ is a fixed damping matrix, $\tilde{T}_e$ is an external force acting on the end-effector, and $\fg$ is our beloved elastic wrench \eqref{eq:geometric_elastic} in $\SE$. Since our task is a PiH, we again fixed the desired configuration $g_d$ as a constant, which leads to the following desired closed-loop system:
\begin{equation} \label{eq:desired_admittance}
    M \dot{V}^b + K_d V^b + \fg(g,g_d) = \tilde{T}_e.
\end{equation}
In what follows, we formulate \eqref{eq:desired_admittance} into a discrete-time setting for the derivation of the admittance control law.
\begin{equation*} 
    V^b(k+1)  = V^b(k) + \Delta t \cdot M^{-1}\left( \tilde{T}_e(k) - \fg(k) - K_d V^b(k) \right),
\end{equation*}
where we use $\fg(k)$ to denote $\fg(g,g_d)$ at $k$ instance for simplicity, and $\Delta t$ is a sampling time. The $V^b(k+1)$ term can then be considered to be the desired velocity at the $k+1$ instance. Thus, the geometric admittance control law now boils down to a joint-space velocity PD control, with desired joint-space velocity $\dot{q}_d(k)$ given as $\dot{q}_d(k) = (J_b(k))^{-1}V^b(k)$.
For the impedance learning problem, the dataset $\{(\eg,(K_p,K_R))_i\}_{i=1}^{N}$ is required. The data collection process is summarized in the training stage of Fig.~\ref{fig:experiment_description}. The robot is provided with $g_d = (R_d, p_d)$ and is controlled by the GAC to execute a PiH task. The human expert supervising this task process changes the gains $(K_p, K_R)$ of GAC in real-time using the GUI, and the gain signals are sent to the manipulator with an ethernet UDP communication protocol. The output signal $\eg$ is received from the communication module and is recorded alongside the gain command signal as the dataset. We collected $75$ trajectories, which sums up to $\sim 250 \mathrm{k}$ of dataset size.
Similar to the simulation experiment, we also collected data and trained policy only on the default case, and tested on the tilted cases, as summarized at the execution stage in Fig.~\ref{fig:experiment_description}.
We set the failure case when the peg is stuck to the whole, e.g., the task cannot be completed within $60\mathrm{s}$.

The result of the hardware validation is presented in Table~\ref{table:BC_result_realrobot}. As can be seen in the result, the proposed approach of utilizing GIC (or GAC) together with the gain scheduling policy implemented with GCEV showed perfect success rates, and the trained policy was transferrable to previously unseen cases. 
\begin{table}
    \setlength\doublerulesep{0.5pt}
    \renewcommand\tabularxcolumn[1]{m{#1}}
    \centering
    \caption{Success rates of the BC policy for the proposed approach in the hardware implementation (Tested 10 times each, Values in Percentage \%)}
    \label{table:BC_result_realrobot}
    \begin{tabularx}{\linewidth}{
     >{\raggedright\arraybackslash\hsize=1.9\hsize}X >{\centering\arraybackslash \hsize=0.7\hsize}X >{\centering\arraybackslash \hsize=0.7\hsize}X >{\centering\arraybackslash \hsize=0.7\hsize}X
    }
    \toprule[1pt]\midrule[0.3pt]
    Method & Default & Case 1 & Case 2 \\
    \midrule
    Proposed (GAC+GCEV) & 100 & 100 & 100 \\
    \midrule[0.3pt]\bottomrule[1pt]
    \end{tabularx}
\end{table}
\vspace{-6pt}
\section{Conclusions and Future Works}
\vspace{-4pt}
In this paper, a geometric approach leveraging $\SE$ group invariance and equivariance for contact-rich robotic manipulation task learning is presented. 
To solve the Peg-in-Hole (PiH) task, the proposed approach builds on top of the geometric impedance control (GIC), where its impedance gains are changed via a left-invariant gain scheduling policy.
Expert behavior cloning is chosen for training the gain scheduling policy. 
Through theoretical analysis, we prove that the proposed GIC and the geometrically consistent error vector (GCEV) used for learning are left-invariant relative to $\SE$ group transformations when represented in the end-effector's body frame system, enabling learning transferability. Furthermore, we show that left invariance in the body-frame representation leads to $\SE$ equivariance of the proposed approach when described in a spatial frame.
A PiH simulation experiment confirms the learning transferability of our proposed method, which is not exhibited by the well-known Cartesian space-based benchmark approach. These results are further validated on an actual PiH robotic hardware implementation, and the pipeline for the hardware implementation is also presented. 

For future work, we will address more realistic scenarios where the exact goal poses are unknown and need to be estimated via sensors, e.g., images or point cloud-based inputs. Moreover, the proposed approach will also be demonstrated in the other types of contact-rich and force-related tasks, such as surface wiping, cable assembly, and pivoting \cite{zhang2023efficient}.







\bibliographystyle{IEEEtran}
\bibliography{references}

\begin{thebibliography}{10}
\providecommand{\url}[1]{#1}
\csname url@samestyle\endcsname
\providecommand{\newblock}{\relax}
\providecommand{\bibinfo}[2]{#2}
\providecommand{\BIBentrySTDinterwordspacing}{\spaceskip=0pt\relax}
\providecommand{\BIBentryALTinterwordstretchfactor}{4}
\providecommand{\BIBentryALTinterwordspacing}{\spaceskip=\fontdimen2\font plus
\BIBentryALTinterwordstretchfactor\fontdimen3\font minus \fontdimen4\font\relax}
\providecommand{\BIBforeignlanguage}[2]{{%
\expandafter\ifx\csname l@#1\endcsname\relax
\typeout{** WARNING: IEEEtran.bst: No hyphenation pattern has been}%
\typeout{** loaded for the language `#1'. Using the pattern for}%
\typeout{** the default language instead.}%
\else
\language=\csname l@#1\endcsname
\fi
#2}}
\providecommand{\BIBdecl}{\relax}
\BIBdecl

\bibitem{ravichandar2020recent}
H.~Ravichandar \emph{et~al.}, ``Recent advances in robot learning from demonstration,'' \emph{Annual review of control, robotics, and autonomous systems}, vol.~3, pp. 297--330, 2020.

\bibitem{zhang2021learning}
X.~Zhang \emph{et~al.}, ``Learning variable impedance control via inverse reinforcement learning for force-related tasks,'' \emph{IEEE Robotics and Automation Letters}, vol.~6, no.~2, pp. 2225--2232, 2021.

\bibitem{beltran2020variable}
C.~C. Beltran-Hernandez \emph{et~al.}, ``Variable compliance control for robotic peg-in-hole assembly: A deep-reinforcement-learning approach,'' \emph{Applied Sciences}, vol.~10, no.~19, p. 6923, 2020.

\bibitem{cohen2016group}
T.~Cohen and M.~Welling, ``Group equivariant convolutional networks,'' in \emph{International conference on machine learning}.\hskip 1em plus 0.5em minus 0.4em\relax PMLR, 2016, pp. 2990--2999.

\bibitem{bekkers2018roto}
E.~J. Bekkers \emph{et~al.}, ``Roto-translation covariant convolutional networks for medical image analysis,'' in \emph{Medical Image Computing and Computer Assisted Intervention--MICCAI 2018: 21st International Conference, Granada, Spain, September 16-20, 2018, Proceedings, Part I}.\hskip 1em plus 0.5em minus 0.4em\relax Springer, 2018, pp. 440--448.

\bibitem{murray1994mathematical}
R.~M. Murray, Z.~Li, and S.~S. Sastry, \emph{A mathematical introduction to robotic manipulation}.\hskip 1em plus 0.5em minus 0.4em\relax CRC press, 1994.

\bibitem{seo2023geometric}
J.~Seo \emph{et~al.}, ``Geometric impedance control on {SE(3)} for robotic manipulators,'' \emph{IFAC World Congress 2023, Yokohama, Japan}, 2023.

\bibitem{zeng2021transporter}
A.~Zeng \emph{et~al.}, ``Transporter networks: Rearranging the visual world for robotic manipulation,'' in \emph{Conference on Robot Learning}.\hskip 1em plus 0.5em minus 0.4em\relax PMLR, 2021, pp. 726--747.

\bibitem{simeonov2022neural}
A.~Simeonov \emph{et~al.}, ``Neural descriptor fields: {SE}(3)-equivariant object representations for manipulation,'' in \emph{2022 International Conference on Robotics and Automation (ICRA)}.\hskip 1em plus 0.5em minus 0.4em\relax IEEE, 2022, pp. 6394--6400.

\bibitem{ryu2022equivariant}
H.~Ryu \emph{et~al.}, ``Equivariant descriptor fields: {SE}(3)-equivariant energy-based models for end-to-end visual robotic manipulation learning,'' in \emph{The Eleventh International Conference on Learning Representations (ICLR)}, 2023.

\bibitem{ryu2023diffusion}
------, ``Diffusion-edfs: Bi-equivariant denoising generative modeling on se (3) for visual robotic manipulation,'' \emph{arXiv preprint arXiv:2309.02685}, 2023.

\bibitem{kim2023robotic}
J.~Kim \emph{et~al.}, ``Robotic manipulation learning with equivariant descriptor fields: Generative modeling, bi-equivariance, steerability, and locality,'' in \emph{RSS 2023 Workshop on Symmetries in Robot Learning}, 2023.

\bibitem{pan2023tax}
C.~Pan \emph{et~al.}, ``Tax-pose: Task-specific cross-pose estimation for robot manipulation,'' in \emph{Conference on Robot Learning}.\hskip 1em plus 0.5em minus 0.4em\relax PMLR, 2023, pp. 1783--1792.

\bibitem{ha2022flingbot}
H.~Ha and S.~Song, ``Flingbot: The unreasonable effectiveness of dynamic manipulation for cloth unfolding,'' in \emph{Conference on Robot Learning}.\hskip 1em plus 0.5em minus 0.4em\relax PMLR, 2022, pp. 24--33.

\bibitem{kim2023se}
S.~Kim \emph{et~al.}, ``{SE(2)}-equivariant pushing dynamics models for tabletop object manipulations,'' in \emph{Conference on Robot Learning}.\hskip 1em plus 0.5em minus 0.4em\relax PMLR, 2023, pp. 427--436.

\bibitem{van2020mdp}
V.~der Pol \emph{et~al.}, ``{MDP} homomorphic networks: Group symmetries in reinforcement learning,'' \emph{Advances in Neural Information Processing Systems}, vol.~33, pp. 4199--4210, 2020.

\bibitem{wang2022equivariant}
D.~Wang \emph{et~al.}, ``Equivariant $ q $ learning in spatial action spaces,'' in \emph{Conference on Robot Learning}.\hskip 1em plus 0.5em minus 0.4em\relax PMLR, 2022, pp. 1713--1723.

\bibitem{wang2022so2}
D.~Wang, R.~Walters, and R.~Platt, ``{SO(2)}-equivariant reinforcement learning,'' \emph{arXiv preprint arXiv:2203.04439}, 2022.

\bibitem{inoue2017deep}
T.~Inoue \emph{et~al.}, ``Deep reinforcement learning for high precision assembly tasks,'' in \emph{2017 IEEE/RSJ International Conference on Intelligent Robots and Systems (IROS)}.\hskip 1em plus 0.5em minus 0.4em\relax IEEE, 2017, pp. 819--825.

\bibitem{kozlovsky2022reinforcement}
S.~Kozlovsky, E.~Newman, and M.~Zacksenhouse, ``Reinforcement learning of impedance policies for peg-in-hole tasks: Role of asymmetric matrices,'' \emph{IEEE Robotics and Automation Letters}, vol.~7, no.~4, pp. 10\,898--10\,905, 2022.

\bibitem{lynch2017modern}
K.~M. Lynch and F.~C. Park, \emph{Modern robotics}.\hskip 1em plus 0.5em minus 0.4em\relax Cambridge University Press, 2017.

\bibitem{khatib1987unified}
O.~Khatib, ``A unified approach for motion and force control of robot manipulators: The operational space formulation,'' \emph{IEEE Journal on Robotics and Automation}, vol.~3, no.~1, pp. 43--53, 1987.

\bibitem{bullo1999tracking}
F.~Bullo and R.~M. Murray, ``Tracking for fully actuated mechanical systems: a geometric framework,'' \emph{Automatica}, vol.~35, no.~1, pp. 17--34, 1999.

\bibitem{lee2010geometric}
T.~Lee \emph{et~al.}, ``Geometric tracking control of a quadrotor uav on {SE}(3),'' in \emph{49th IEEE conference on decision and control (CDC)}.\hskip 1em plus 0.5em minus 0.4em\relax IEEE, 2010, pp. 5420--5425.

\bibitem{caccavale1999six}
F.~Caccavale \emph{et~al.}, ``Six-dof impedance control based on angle/axis representations,'' \emph{IEEE Transactions on Robotics and Automation}, vol.~15, no.~2, pp. 289--300, 1999.

\bibitem{robosuite2020}
Y.~Zhu \emph{et~al.}, ``robosuite: A modular simulation framework and benchmark for robot learning,'' in \emph{arXiv preprint arXiv:2009.12293}, 2020.

\bibitem{ochoa2021impedance}
H.~Ochoa and R.~Cortes{\~a}o, ``Impedance control architecture for robotic-assisted mold polishing based on human demonstration,'' \emph{IEEE Transactions on Industrial Electronics}, vol.~69, no.~4, pp. 3822--3830, 2021.

\bibitem{shaw2022rmps}
S.~Shaw, B.~Abbatematteo, and G.~Konidaris, ``{RMPs} for safe impedance control in contact-rich manipulation,'' in \emph{2022 International Conference on Robotics and Automation (ICRA)}.\hskip 1em plus 0.5em minus 0.4em\relax IEEE, 2022, pp. 2707--2713.

\bibitem{todorov2012mujoco}
E.~Todorov, T.~Erez, and Y.~Tassa, ``Mujoco: A physics engine for model-based control,'' in \emph{2012 IEEE/RSJ International Conference on Intelligent Robots and Systems}.\hskip 1em plus 0.5em minus 0.4em\relax IEEE, 2012, pp. 5026--5033.

\bibitem{BerkeleyRLkit}
``{Berkeley RL Kit},'' \url{https://github.com/rail-berkeley/rlkit}, accessed: 2023-07-01.

\bibitem{tobin2017domain}
J.~Tobin \emph{et~al.}, ``Domain randomization for transferring deep neural networks from simulation to the real world,'' in \emph{2017 IEEE/RSJ international conference on intelligent robots and systems (IROS)}.\hskip 1em plus 0.5em minus 0.4em\relax IEEE, 2017, pp. 23--30.

\bibitem{zhang2023efficient}
X.~Zhang \emph{et~al.}, ``Efficient sim-to-real transfer of contact-rich manipulation skills with online admittance residual learning,'' \emph{arXiv preprint arXiv:2310.10509}, 2023.

\end{thebibliography}
\end{document}